\documentclass{article}
\usepackage[ruled,vlined,linesnumbered]{algorithm2e}
\usepackage{amsmath} 
\usepackage{graphicx}
\usepackage{float}
\usepackage{caption}
\usepackage{subcaption}
\usepackage{diagbox}
\usepackage{wrapfig}
\usepackage{booktabs}
\usepackage{titlesec}
\usepackage[preprint]{corl_2025} 
\usepackage{amssymb}
\let\oldnl\nl
\newcommand{\nonl}{\renewcommand{\nl}{\let\nl\oldnl}}

\titlespacing{\section}{0pt}{4pt}{1pt}
\titlespacing{\subsection}{0pt}{3pt}{1pt}
\titlespacing{\subsubsection}{0pt}{2pt}{0.5pt}

\newcounter{myenum}
{\end{list}}

\newenvironment{flushitemize}{%
\begin{list}{$\bullet$}
   {\setlength{\leftmargin}{15pt}}%
    \setlength{\labelwidth}{20pt}
    \setlength{\itemindent}{0pt}
    \setlength{\labelsep}{0.5em}
 \setlength{\itemsep}{1pt}
 \setlength{\parskip}{0pt}
 \setlength{\parsep}{0pt}}
{\end{list}}

\title{AsymDex: Asymmetry and Relative Coordinates \\ for RL-based Bimanual Dexterity}

\author{
    Zhaodong Yang$^1$, Yunhai Han$^1$, Ai-Ping Hu$^1$, Harish Ravichandar$^1$ \\
    $^1$Georgia Institute of Technology \\
    \texttt{\{halyang, yhan389, ahu6,  harish.ravichandar\}@gatech.edu} \\
}
\begin{document}
\maketitle


\begin{abstract}
We present \textit{Asymmetric Dexterity (AsymDex)}, a novel and simple reinforcement learning (RL) framework that can efficiently learn a large class of bimanual skills in multi-fingered hands without relying on demonstrations. Two crucial insights enable AsymDex to reduce the observation and action space dimensions and improve sample efficiency. 
First, true ambidexterity is rare in humans and most of us exhibit strong ``handedness".
Inspired by this observation, we assign complementary roles to each hand: the \textit{facilitating hand} repositions and reorients one object, while the \textit{dominant hand} performs complex manipulations to achieve the desired result (e.g., opening a bottle cap, or pouring liquids). Second, controlling the \textit{relative} motion between the hands is crucial for coordination and synchronization of the two hands. As such, we design relative observation and action spaces and leverage a relative-pose tracking controller. 
Further, we propose a two-phase decomposition in which AsymDex can be readily integrated with recent advances in grasp learning to facilitate both the acquisition and manipulation of objects using two hands. 
Unlike existing RL-based methods for bimanual dexterity with multi-fingered hands, which are  either sample inefficient or tailored to a specific task, AsymDex can efficiently learn a wide variety of bimanual skills that exhibit asymmetry. Detailed experiments on seven asymmetric bimanual dexterous manipulation tasks (four simulated and three real-world) reveal that AsymDex consistently outperforms strong baselines that challenge our design choices.
The project website is at \url{https://sites.google.com/view/asymdex-2025/}.
\end{abstract}



\section{Introduction}
\label{sec:intro}
Bimanual dexterity is crucial for robots operating in human environments as they allow for complex yet flexible manipulation compared to a single hand~\cite{smith2012dual,grannen2023stabilize, avigal2022speedfolding, chi2024universal, wang2024dexcap, zakka2023robopianist}. We are interested in learning a wide range of bimanual dexterous skills on multi-fingered hands purely from reinforcement. 

While learning on a single multi-fingered hand is known to be challenging~\cite{Rajeswaran2018DAPG, OpenAI2018Cube, qi2022hand, han2023utility, han2024learning, chenkorol, shaw2024learning}, learning bimanual dexterity is made more challenging due to the higher-dimensionality and the need for coordination and synchronization of two hands~\cite{smith2012dual,chen2022towards}. These challenges are only exacerbated when these skills have to be learned from reinforcement, explaining why existing efforts often either resort to expert demonstrations~\cite{chen2024object, qiu2025humanoid, jiang2024dexmimicgen, zhou2024learning} or limit themselves to specific tasks~\cite{lin2024twisting, huang2023dynamic}. 

We rely on two insights to tackle the challenges of bimanual dexterity.
First, we are inspired by how humans and other great apes approach these challenges: there is a natural \textit{asymmetry} in how we use each of our hands when we perform most bimanual tasks~\cite{guiard1987asymmetric}. Specifically, we tend to use one hand to reposition and reorient an object so as to make it easier for the other hand to perform complex manipulation. While leveraging such asymmetry might appear to restrict the class of bimanual skills we can learn, rich bodies of work in human biomechanics and evolution reveal its prevalence and necessity~\cite{guiard1987asymmetric,kimmerle2010development, sainburg2002evidence, studenka2008influence}. Evolutionary biologists posit handedness evolved to meet the escalating cognitive demands of tool use, language, and complex manipulation~\cite{cashmore2008evolution}. Indeed, a large number of real-world tasks admit such asymmetry (e.g., attachment, detachment, assembly, and pouring).

Looking closely at the asymmetry in bimanual dexterity reveals our second insight. Our non-dominant hand tends to hold an object firmly as we reorient and reposition it \textit{relative} to the dominant hand or the object being held by the dominant hand (e.g., tilting a pen before uncapping). This suggests that there is often little to no in-hand movement of the object grasped by the non-dominant hand, and robust synchronization can be achieved by ensuring relative movement of the two hands.

\begin{figure}[t]
    \centering
    \includegraphics[width=\textwidth]{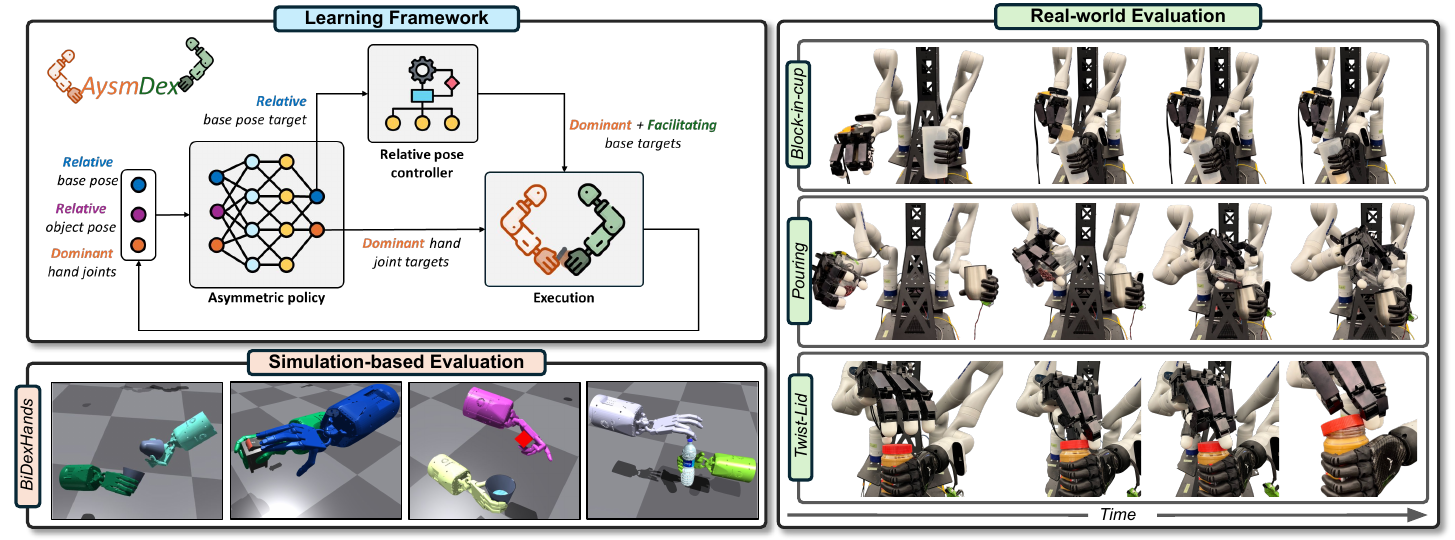}
    \caption{Our approach (AsymDex) efficiently learns asymmetric bimanual dexterous manipulation skills based on reinforcement learning by effectively leveraging i) the natural asymmetry in the hands' roles and ii) relative state and action spaces that prioritize synchronization.}
    \label{fig:overview}
\end{figure}

We contribute a novel RL-based learning framework for bimanual dexterity, dubbed \textit{Asymmetric Dexterity (AsymDex)} by operationalizing the above two insights (see Fig.~\ref{fig:block_diagram}).
To incorporate asymmetry, we define a \textit{dominant hand} and a \textit{facilitating hand}. While the dominant hand learns complex skills across all its degrees of freedom, the facilitating hand learns to reposition and reorient the object by controlling the 6D pose of its base (i.e., no finger movement). This allows us to both reduce the dimensionality of the observation and action spaces and tightly integrate the roles of the two hands.
To ensure coordination and synchronization, AsymDex operates over relative observation and action spaces that incentivize flexible coordination of the two hands without resorting to explicit time-dependence or task-specific coordinate frame designs. 

We also leverage the observation that bimanual manipulation in practice is composed of two distinct phases: i) the \textit{acquisition phase} in which objects are grasped, and ii) the \textit{interaction phase} in which the two hands coordinate to perform the bimanual task. Unlike many existing methods that entirely ignore the acquisition phase\cite{grannen2023stabilize, chen2022towards, lin2024twisting, bahety2024screwmimic}, we show that this decomposition enables AysmDex to be seamlessly integrated with learned grasping policies to enable fluent execution. 


In summary, we contribute AsymDex -- a novel Rl-based framework for learning a wide variety of bimanual dexterous skills by taking inspiration from two key aspects of human bimanual dexterity: i) asymmetric hand roles, and ii) relative hand movement. 
We conduct comprehensive experiments on seven complex tasks (four simulated and three real) and compare against strong baselines that challenge the need for AsymDex's structural inductive biases. Our results show that AsymDex consistently outperforms these baselines in terms of both task performance and sample efficiency.

\section{Related Work}
\label{sec:related_work}

\textbf{Learning Bimanual Manipulation}: 
Several existing methods focus on learning bimanual skills, but are often limited to simple end-effectors.
Imitation learning (IL) based approaches have been particularly successful in bimanual manipulation~\cite{avigal2022speedfolding, franzese2023interactive, seo2023deep, bahety2024screwmimic}, and have led to novel and low-cost infrastructure to collect bimanual manipulation data~\cite{chi2024universal,zhao2023learning}. These approaches rely on demonstrations to provide the necessary supervision to learn effective coordination strategies.
Reinforcement learning (RL) has also been shown to be successful in learning bimanual manipulation skills~\cite{lin2023bi, kataoka2022bi, chitnis2020efficient,li2023efficient}. These methods implicitly incentivize coordination by learning to optimize reward functions that favor task success and efficiency. 
In contrast to all of these works that only consider parallel jaw grippers, AsymDex learns bimanual dexterous manipulation skills involving multi-fingered hands.

\textbf{Asymmetry in Bimanual Manipulation}: Motivated by the asymmetry in how humans use their two hands (referred to as \textit{role-differentiated bimanual manipulation}~\cite{kimmerle2010development, sainburg2002evidence, studenka2008influence, jo2020analysis}), recent works assign different roles to each robot hand in the bimanual system~\cite{grannen2023stabilize, holladay2024robust, grannen2022learning,liu2022robot, cui2024task}. However, some of these approaches restrict the role of the facilitating hand to stabilizing the object while the dominant hand manipulates it~\cite{grannen2023stabilize, holladay2024robust, grannen2022learning}. In contrast, AsymDex allows the facilitating hand to reposition and reorient the object \textit{simultaneously} as the dominant hand executes its role. 
Importantly, unlike AsymDex, all these prior methods are limited to parallel jaw grippers.

\textbf{Learning Dexterous Manipulation}:
Learning dexterous manipulation skills involves addressing numerous challenges due to high dimensional state and action spaces and highly nonlinear dynamics. Recent works have tackled these challenges using imitation learning (IL) or reinforcement learning (RL) and demonstrate impressive performance~\cite{Rajeswaran2018DAPG, OpenAI2018Cube, qi2022hand, han2023utility, han2024learning, shaw2024learning, nagabandi2020deep, qin2022dexmv, Khandate-RSS-23, xie2023neural}. However, IL-based methods rely either on complex infrastructure and retargeting methods to collect demonstrations ~\cite{Rajeswaran2018DAPG, shaw2024learning, qin2022dexmv, handa2020dexpilot, arunachalam2023holo} or pre-trained expert policies~\cite{han2023utility, han2024learning, xie2023neural}. On the other hand, RL-based methods do not share these constraints as they learn skills via reinforcement, but tend to require significant exploration even for dexterous manipulation with a single hand~\cite{OpenAI2018Cube, qi2022hand, nagabandi2020deep, Khandate-RSS-23}. 
As we show in our experiments, naive application of RL-based methods is not effective for bimanual dexterous manipulation due to the increased dimensionality and the need for coordination. 

\textbf{Learning Bimanual Dexterous Manipulation}: A few recent studies have focused on learning bimanual dexterity. 
Some of these methods require the collection of expert demonstrations~\cite{wang2024dexcap} and suffer from the same limitations we discussed earlier for IL-based methods that use parallel jaw grippers.
To circumvent the need for collecting demonstrations, recent efforts have led to methods that only leverage RL and yet are capable of learning impressive bimanual manipulation skills, such as playing the piano~\cite{zakka2023robopianist}, twisting lids off containers~\cite{lin2024twisting}, and dynamic handover~\cite{huang2023dynamic}. While these methods are specifically designed to solve a particular task, AsymDex is capable of efficiently learning different bimanual dexterous manipulation tasks. Some recent studies investigate generalized learning method by utilizing expert demonstration for efficient RL training \cite{chen2024object, qiu2025humanoid, jiang2024dexmimicgen, zhou2024learning}, while AsymDex can efficiently learn bimanual manipulation skills without relying on demonstration.

\section{Problem Formulation}
\label{sec:problem}
We first formulate the general problem of bimanual dexterous manipulation, and then introduce asymmetric bimanual dexterity.

Consider the general problem of bimanual dexterous manipulation, in which two multi-fingered hands coordinate to manipulate up to two objects. Formally, this problem can be defined as a Partially-Observable Markov Decision Process (POMDP) $\mathcal{M}=(\mathcal{S}, \mathcal{Z}, \mathcal{A}, \mathcal{R}, \mathcal{P})$, where $\mathcal{S}\in \mathbb{R}^n$ is the state space, $\mathcal{Z} \in \mathbb{R}^m$ is the observation space, $\mathcal{A} \in \mathbb{R}^u$ is the action space, ${\mathcal{R}: \mathbb{R}^m \times \mathbb{R}^u \rightarrow \mathbb{R}}$ is the reward function, and ${\mathcal{P}: \mathbb{R}^n \times \mathbb{R}^u \rightarrow \mathbb{R}^n}$ is the environment dynamics. Note that we do not assume access to any demonstrations. Instead, we tackle of challenge of learning purely based on reinforcement. Given this formulation, the problem boils down to learning a policy $\pi: \mathcal{Z} \rightarrow \mathcal{A}$ that maximizes the expected discounted cumulative reward $E_{\pi}[\Sigma_{t=0}^{T-1}\gamma^t \mathcal{R}(z(t), a(t))]$. 


\textbf{Observation and Action Spaces}: The observation space $\mathcal{Z}$ is composed of hand and object states. At time step $t$, $z(t)=[\xi_1(t),\xi_2(t),\xi^{obj}(t)]$, where  $\xi_1(t)$ contains the first hand's current full (fingers + wrist) configuration $\xi^h_1(t) \in \mathbb{R}^{n_1}$ and the 6 DOF pose of its base $\xi^b_1(t) \in \mathbb{SE}(3)$,  
$\xi_2(t)$ contains the corresponding elements for the second hand, 
and $\xi^{obj}(t)$ contains the 6 DOF poses of either two objects (e.g., stacking two cups) or parts of one object (e.g., bottle and lid).
The joint action at time step $t$ is given by $a(t)=[\hat{\xi}_1(t),\hat{\xi}_2(t)]$, where $\hat{\xi}_1(t)$ denotes the target joint configuration of the first hand $\hat{\xi}^h_1(t)$ and the target 6 DOF pose of its base $\hat{\xi}^b_1(t)$, and $\hat{\xi}_2(t)$ denotes the corresponding targets for the second hand. With the above definitions, we can now define the problem of learning bimanual dexterity as one of learning a monolithic symmetric policy: 
$\pi_{\mathrm{sym}}(a(t) | z(t))$.

\textbf{Asymmetric Dexterity Problem}:
Asymmetric dexterity can be viewed as a broad subclass to the above general class of problems. Inspired by strategies employed by humans and other great apes, we are interested in tasks in which one multi-fingered hand performs complex and precise manipulations while the other plays a facilitating role by supporting and actively reorienting objects of interest (e.g., stacking, attachment, detachment, etc.). Note that our formulation does not restrict the movement of the second hand; it merely restricts the relative movement between the second hand and the object being held. As explained in Sec.~\ref{sec:intro}, a large number of bimanual tasks exhibit asymmetry, hinting at handedness in most humans.  
We are thus interested in learning an asymmetric policy $\pi_{\mathrm{AsymDex}}(\cdot)$ with carefully-defined observation $\mathcal{Z}_{\mathrm{AsymDex}}$ and action $\mathcal{A}_{\mathrm{AsymDex}}$ spaces in an effort to improve both effectiveness and sample efficiency.

Note that our primary contributions and the asymmetric assumption pertain to the \textit{interaction phase} of bimanual dexterity, in which the two hands actively coordinate to complete the task after having grasped the necessary object(s). Most existing works focus solely on the interaction phase~\cite{lin2024twisting, huang2023dynamic, chen2023vegetable}. In Sec.~\ref{sec:acquiring}, we discuss how our approach can be readily extended to also tackle the \textit{acquisition phase} (learning to grasp the necessary objects before coordinating).

\section{AsymDex: Learning Asymmetric Dexterity}
\label{sec:method}

While one could learn the symmetric policy $\pi_{sym}(\cdot)$ as defined in Sec.~\ref{sec:problem}, training such a policy can be inefficient or ineffective due to the high dimensionality of the observation and action spaces (see Sec.~\ref{sec:evaluation} for empirical evidence). Importantly, a symmetric approach ignores the natural asymmetry found in most bimanual tasks. Below, we explain how AsymDex overcomes these challenges.

\subsection{Incorporating Asymmetry}
\label{sec:asym_only}


Motivated by the natural asymmetry in human bimanual manipulation~\cite{kimmerle2010development, sainburg2002evidence, studenka2008influence}, we assign different roles to each robot hand during their interaction: a \textit{facilitating hand} that is responsible for holding and repositioning and reorienting the object, and a \textit{dominant hand} which is responsible for fine-grained dexterous manipulation of the object(s).
We note that there tends to be no relative motion between the facilitating hand and the grasped object in asymmetric dexterity since the facilitating hand need only hold, move, and reorient the object (i.e., no in-hand reorientation). In contrast, the dominant hand can interact freely with the object(s). This suggests that the asymmetric dexterity is neither dependent on nor influences the finger joints of the facilitating hand during the interaction phase. As such, we can considerably reduce the observation and action spaces by defining an asymmetry-only bimanual policy: $\pi_{\mathrm{asym}}(a_{\mathrm{asym}}(t) | z_{\mathrm{asym}}(t))$ with actions $a_{\mathrm{asym}}(t)=[\hat{\xi}_d(t), \hat{\xi}^b_f(t)]$ and observations $z_{\mathrm{asym}}(t)=[\xi_d(t), \xi^b_f(t), \xi^{obj}(t)]$. This change both reduces the dimensionality and ensures that the facilitating hand doesn't learn unproductive or unnecessary behaviors.

\subsection{Incorporating Relative Observation and Action Spaces}
\label{sec:ours}
\begin{wrapfigure}{r}{0.32\textwidth}
\vspace{-2.5em}
\resizebox{0.32\textwidth}{!}{%
\centering
    \includegraphics[width=\textwidth]{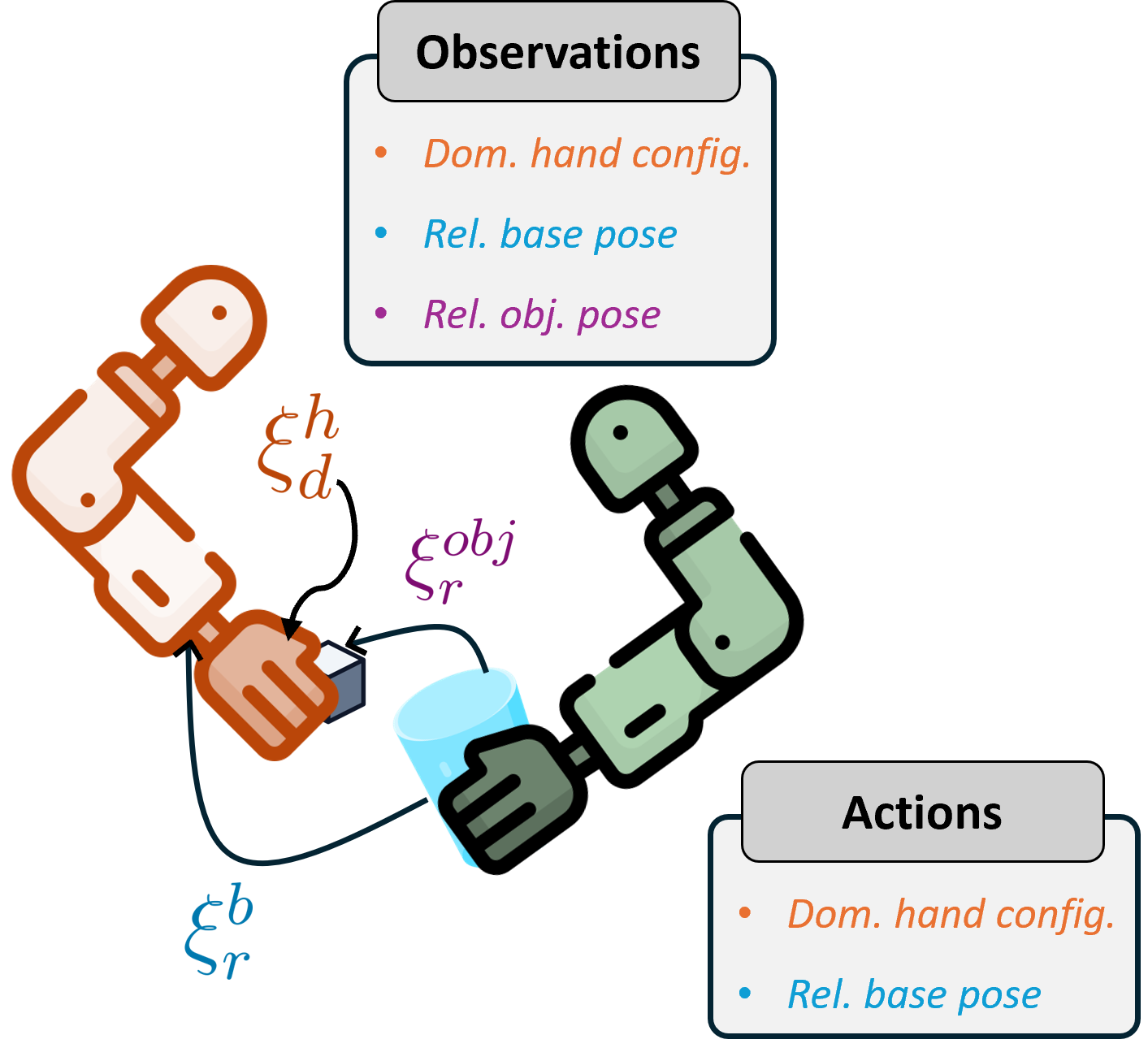}
}
\caption{\small{AsymDex's observation and Action Spaces.}}\label{tab:two real world}
\vspace{-1.5em}
\end{wrapfigure} 
In addition to asymmetry, a key characteristic of bimanual dexterity is the synchronized and responsive movement of the two hands. We can further reduce the size of the observation and action spaces and introduce tight coupling between the hands' behaviors by defining \textit{relative} and \textit{object-centric} coordinates that capture the relationships between the movements of two hands and the object(s) being manipulated. 
Indeed, the use of relative state spaces has shown to considerably benefit bimanual manipulation with simple end effectors~\cite{bahety2024screwmimic,laha2021coordinated, chiacchio1996direct, tarbouriech2018dual}. While some of these prior works limit the relative space to a one-degree-of-freedom (1-DoF) action space~\cite{bahety2024screwmimic}, AsymDex allows for complete 6-DoFs relative space.

Let $\xi^{obj}_f(t)$ be the state of the object being held by the facilitating hand, and let $\xi^{obj}_d(t)$ be the state of the object being manipulated by the dominant hand. We attach a coordinate frame to the object being held by the facilitating hand: $P_{f}$. Now, we can transform the asymmetry-only observations $z_{\mathrm{asym}}(t)$ (originally defined in the absolute or world coordinate frame) into the new coordinate frame $P_f$. Note that since there is no relative motion between the facilitating hand and the object that it is holding, neither $\xi^{obj}_f(t)$ nor $\xi^b_f(t)$ change in $P_f$, and can thus be removed from our observation space without losing any information. Now, transforming the observations ($\xi^b_d(t), \xi^{obj}_d(t)$) onto $P_f$ yields $z_{\mathrm{AsymDex}}=[\xi^h_d(t), \xi^b_r(t), \xi^{obj}_r(t)]$, where $\xi^b_r(t)$ and $\xi^{obj}_r(t)$ respective denote the 6D relative poses of the dominant hand base and the object being manipulated by the dominant hand, both defined with respect to the object being held by the facilitating hand. Note that since $\xi^h_d(t)$ denotes the dominant hands' joint states, it is not impacted by the change of coordinates.
Similarly, we apply the same modifications to the asymmetry-only action $a_{asym}(t)$, yielding $a_{\mathrm{AsymDex}}(t)=(\hat{\xi}^b_r(t), \hat{\xi}^h_d(t))$, where $\hat{\xi}^b_r(t)$ is the target relative pose of the dominant hand now defined relative to the object being held by the facilitating hand. 
Incorporating the above change of coordinates in addition to leveraging asymmetry, allows us to define AsymDex's policy as  $\pi_{\mathrm{AsymDex}}(a_{\mathrm{AsymDex}}(t) | z_{\mathrm{AsymDex}}(t))$. Note that our formulation has significantly reduced the dimensions of both the state and action spaces, compared to the symmetric policy $\pi_{\mathrm{sym}}$ as defined in Section~\ref{sec:problem}. 

We parameterize the AsymDex policy $\pi_{\mathrm{AsymDex}}(\cdot)$ using an MLP and using Proximal Policy Optimization (PPO)~\cite{schulman2017proximal} to train it. See Appendix. A 
for details of the algorithm and policy architecture.

\subsection{Relative Pose Tracking Controller}
\label{sec:rel_controller}
To control the hand bases based on the target relative pose $\hat{\xi}^b_r(t)$ provided by $\pi_{\mathrm{AsymDex}}$, we designed a bimanual controller that computes both the target dominant hand base pose $\hat{\xi}^b_d(t)$ and the target facilitating hand base pose $\hat{\xi}^b_f(t)$ as follows
\begin{equation}
\label{eqn:controller}
\begin{split}
\hat{\xi}^b_d(t) & = \alpha R^{o_f}_{world}\cdot dist(\hat{\xi}^b_r(t), \xi^b_r(t)) + \xi^b_d(t), \\
\hat{\xi}^b_f(t) & = (\alpha - 1) R^{o_f}_{world}\cdot dist(\hat{\xi}^b_r(t), \xi^b_r(t)) + \xi^b_f(t),   
\end{split}
\end{equation}
where $R^{o_f}_{world}$ denotes the rotational transformation from Frame $P_{f}$ to the world frame $P_W$, $dist(\cdot)$ denotes the difference between two 6D poses, and $\alpha$ is a hyperparameter that controls the relative involvement of each hand. 
The pseudo-code of the training process is included in Alg. 1. 

\subsection{Acquisition Phase}
\label{sec:acquiring}
\begin{figure}[t]
    \centering
    \includegraphics[width=\textwidth]{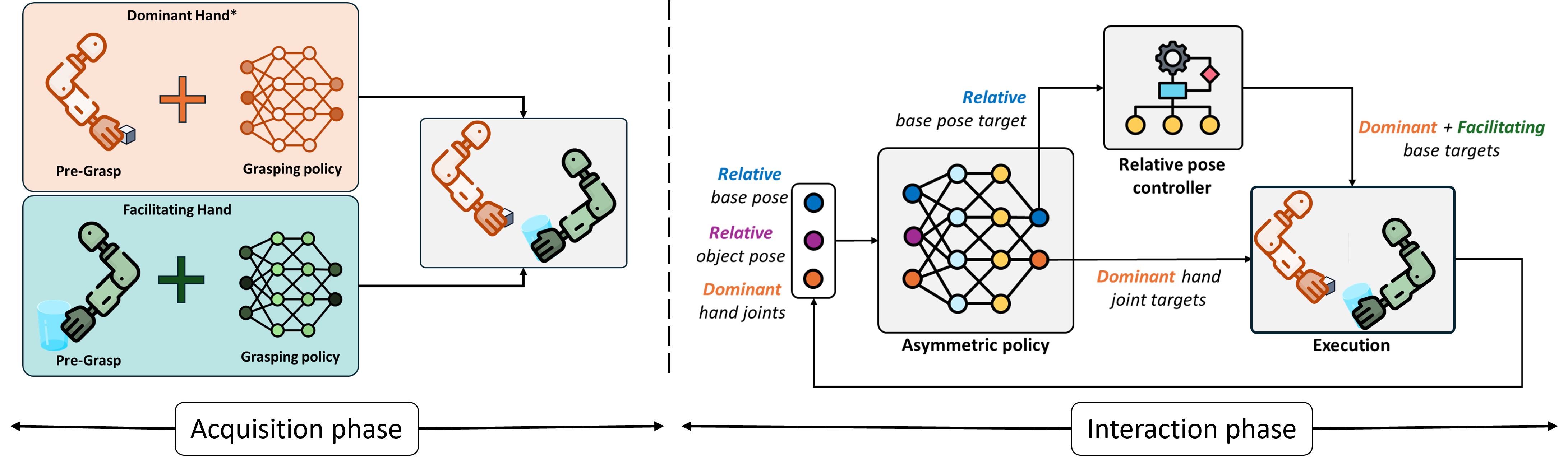}
    \caption{We decompose asymmetric bimanual dexterous manipulation into two phases: An \textit{acquisition} phase and an \textit{interaction} phase. We show that AsymDex can be readily integrated with learned grasping policies in order to seamlessly acquire and manipulate objects.}
    \label{fig:block_diagram}
\end{figure}

While our approach as explained thus far deals with the challenge of coordinating two hands to accomplish asymmetric dexterous manipulation tasks, it assumes that the object(s) of interest have already been grasped at the beginning of the task. However, in practice, robots must be learn to grasp and pick up the necessary objects before the interaction between the two hands (and the objects) can begin. We refer to this initial phase as the \textit{acquisition} phase.
Most recent works on bimanual dexterous manipulation often entirely ignore the acquisition phase and focus purely on the interaction phase~\cite{zakka2023robopianist, lin2024twisting, huang2023dynamic}. 
In contrast, we demonstrate that our approach can seamlessly accommodate the acquisition phase by i) leveraging the observation that the acquisition phase doesn't require the coordination of two arms, and ii) employing recent advances in learning to grasp.
Specifically, we demonstrate that we can seamlessly integrate AsymDex with PDGM~\cite{dasari2023pgdm},  which can efficiently learn multi-fingered grasping policies by leveraging pre-grasp poses (see Fig.~\ref{fig:block_diagram}). Details about the grasping reward design are available in Appendix. B. 
We begin by executing the grasping policy in isolation and then "turn on" the asymmetric policy learned by AsymDex after the object has been firmly grasped by the facilitating hand.
If the task requires the dominant hand to also grasp a second object, we employ the same method to train a grasping policy for dominant hand to acquire the object, but switch the control of the dominant hand's joints over the asymmetric policy after the object has been grasped. 

\section{Experimental evaluation}
\label{sec:evaluation}

We evaluated AsymDex on four simulated and three real-world bimanual dexterous tasks and compared its performance against strong baselines that challenge our key design choices. 

\subsection{Simulation Experiments}
Our experiments in simulation both systematically and rigorously evaluate AsymDex.

\textbf{Tasks}:  
We evaluated AsymDex and the baselines on the following four bimanual manipulation tasks which contain both original (\textit{Block in cup}, \textit{Bottle cap}) and adapted tasks (\textit{Stack}, \textit{Switch}) from BiDexHand~\cite{chen2022towards} (see Fig.~\ref{fig:overview}). All these tasks use two Shadow Hands -- each a 30-DoF simulated multi-finger hand system (24-DoF hand + 6-DoF floating wrist base) built with Isaac Gym~\citep{liang2018gpuaccelerated}.

\begin{flushitemize}
\item \textit{Block in cup}: 
The two hands must coordinate to ensure that one hand places a block inside a cup that is being held by the other without letting either the cup or the block fall to the ground. 
\item \textit{Stack}: 
Two cups need to be stacked together. Each hand must hold a cup, and both must coordinate such that the two cups are aligned as one slides into the other.
\item \textit{Bottle cap}: 
One hand must hold and reorient a bottle such that the other hand can grasp and separate the bottle cap from the bottle.
\item \textit{Switch}: 
One hand holds and reorients a switch in a way that allows the other hand to turn it on.
\end{flushitemize}

Note that our task designs are more challenging than their counterparts in BiDexHand~\cite{chen2022towards}. We require that the two hands coordinate and synchronize to achieve success in each of the above four tasks, especially since (unlike the original designs) we do not provide a support surface (e.g., a table) that would significantly reduce the need for bimanual coordination. See Appendix. C
for details on state space design, sampling procedure, success criteria, and reward design.




\textbf{Metrics}: 
We quantify performance in terms of i) \textit{success rate} (see Appendix. C
for criteria) and ii) \textit{sample efficiency}. We report both metrics across five random seeds in all experiments.



\subsubsection{Learning Bimanual Coordination}
We first evaluated AsymDex's effectiveness during the interaction phase. Following contemporary practice in methods that learn bimanual dexterous skills~\cite{lin2024twisting,huang2023dynamic}, we initialized the environment such that the hands are at appropriate pre-grasp poses near the respective objects. 
This allows us to isolate and examine AsymDex's ability to learn to coordinate two multi-fingered hands. See Section~\ref{experiment:two stage} for the second experiment in which we also consider the challenge of acquiring the objects from a tabletop surface before interaction begins.

We compared AsymDex against the following baselines:
\begin{flushitemize}
\item \texttt{Sym}: This policy assumes that both hands play an equal role in bimanual manipulation (see $\pi_{sym}$ in Sec.~\ref{sec:problem}). This baseline allows us to examine the necessity and effectiveness of leveraging the asymmetry in hand roles as well as the relative action and observation spaces.
\item \texttt{Asym-w/o-rel}: This policy leverages asymmetry in hand roles, but learns over absolute observation and action places (see Sec.~\ref{sec:asym_only}). As such, this baseline allows us to examine the necessity and effectiveness of relative action and observation spaces. 
\item \texttt{Rel-w/o-asym}: This policy leverages the relative observation and action places, but ignores asymmetry. As such, it allows us to examine the necessity and effectiveness of asymmetry.
\end{flushitemize}

\begin{figure}[b]
    \centering
     \subfloat[BiDexHand task training curve]{\includegraphics[width=0.55\textwidth]{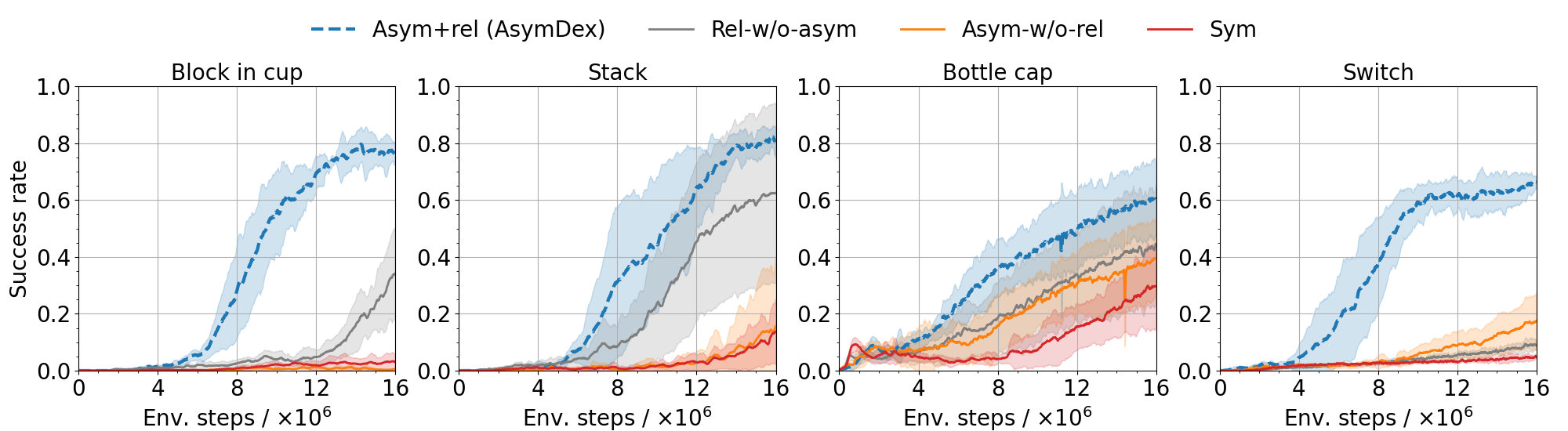}}
     \hfill
     \subfloat[Real-world task training curve]{\includegraphics[width=0.45\textwidth]{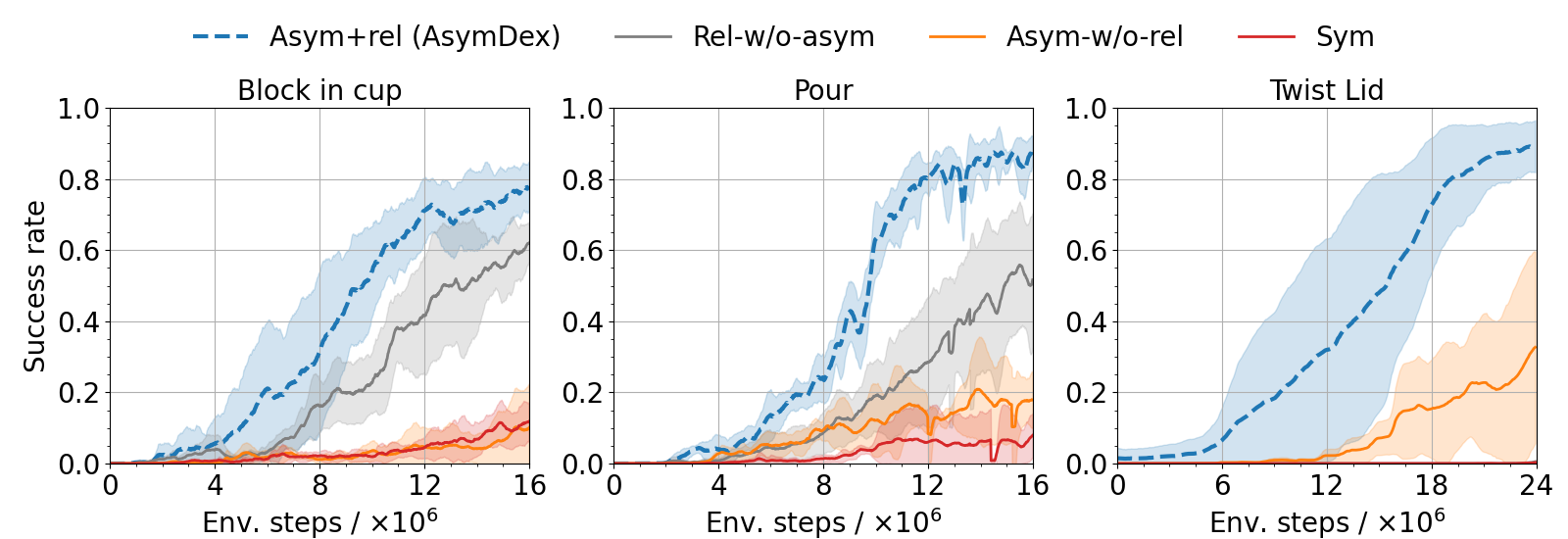}}
    \caption{AsymDex consistently outperforms the baselines in terms of sample efficiency and success rate. Solid lines indicate mean trends and shaded areas show $\pm$ std. dev., over five random seeds. }
    \label{fig:learning curve}
\end{figure}



\textbf{Both asymmetry and relative spaces are necessary for consistent performance}: We report the learning curves in Fig.~\ref{fig:learning curve} (a) and success rates in Table~\ref{tab:interaction phase}. Note that AsymDex consistently outperforms all the baselines across all four tasks in terms of success rate and sample efficiency, with significant margins in two tasks (\textit{Block in cup} and \textit{Switch}).
\texttt{Rel-w/o-asym} performs better than the other two baselines across all tasks except \textit{Switch},
while \texttt{Asym-w/o-rel} performs better than \texttt{Monolithic} on all tasks except \textit{Block in cup}, on which both struggle.

Taken together, the above observations reveal a few key insights. 
First, when used in isolation, neither asymmetry nor relative spaces are sufficient across all tasks. 
Second, the use of relative spaces offers a larger boost in performance compared to asymmetry, likely due to the fact that relative spaces avoid unnecessary exploration (e.g., when the two hands move in parallel) while allowing the facilitating hand to exhibit more complex behaviors. Third, ignoring both asymmetry and relative spaces (\texttt{Sym}) hardly leads to success. 

\begin{table}[t]
    \centering
    \begin{tabular}{c|c|c|c|c}
    \hline
   \diagbox[width=1in]{\textbf{Task}}{\textbf{Method}} & \texttt{Sym}    & \texttt{Asym-w/o-rel} & \texttt{Rel-w/o-Asym} & \texttt{AsymDex} (ours) \\
        \hline
    \textit{Block in cup} & $ 0.0429 \pm 0.0266 $ &  $ 0.0164 \pm 0.0190 $ & $ 0.1086 \pm 0.1378$  & \textbf{ 0.7701 $\pm$ 0.0559} \\
        \hline
    \textit{Stack} & $ 0.0771 \pm 0.0611 $ &  $ 0.2185 \pm 0.3232 $  & $ 0.6560 \pm 0.3213 $ &   \textbf{0.8392 $\pm$ 0.0596}   \\
        \hline
    \textit{Bottle cap} & $ 0.3111 \pm 0.1813 $ &  $ 0.4143 \pm 0.1410 $  & $ 0.4730 \pm 0.2011$ & \textbf{ 0.6295 $\pm$ 0.1422} \\
        \hline
    \textit{Switch}  & $ 0.0563 \pm 0.0126 $ &  $ 0.1626 \pm 0.0882 $  & $ 0.1149 \pm 0.0176 $ & \textbf{ 0.6700 $\pm$ 0.0359} \\
        \hline
    \end{tabular}
    \caption{Success rates (mean $\pm$ std. dev.) for the interaction phase.}
    \label{tab:interaction phase}
\end{table}

\subsubsection{Learning to Grasp and Coordinate}
\label{experiment:two stage}
We next evaluated AsymDex's ability to incorporate the object acquisition phase before the interaction phase. Specifically, we initialize the environment for each task such that the objects of interest are placed on a tabletop surface. As such, each method needs to learn both to grasp the necessary objects and to coordinate the two hands to complete the tasks. 


\begin{table}[b]
    \caption{Success rates (mean $\pm$ std. dev.) after combining acquisition and interaction phases}
    \centering
    \begin{tabular}{c|c|c|c}
    \hline
   \diagbox[width=1in]{\textbf{Task}}{\textbf{Method}} & \texttt{Monolithic}    & \texttt{2-stage-sym} & \texttt{AsymDex} \\
        \hline
    \textit{Block in cup} & $0.0321 \pm 0.0251$ &  $0.1505 \pm 0.1059$  & \textbf{0.7938 $\pm$ 0.0897} \\
        \hline
    \textit{Bottle cap} & $0.0 \pm 0.0$ &  $0.2552 \pm 0.1573$  & \textbf{0.6116 $\pm$ 0.1328} \\
        \hline
    \end{tabular}
    \label{tab:two stage}
\end{table}

For AsymDex, we follow the same strategy introduced in Section~\ref{sec:acquiring}, and comapare its performance against the following baselines:
\begin{flushitemize}
\item \texttt{Monolithic}: 
This baseline uses a single policy to learn both the grasping and interaction phases for both hands, allowing us to investigate the benefits of two-phase decomposition. 
\item \texttt{2-stage-sym}: 
This policy benefits from the two phase decomposition but leverages neither asymmetry nor relative spaces. As such, this baseline allows us to examine if merely employing two-phase decomposition is sufficient.
\end{flushitemize}
To ensure a fair comparison, we provide pre-grasp pose annotations to both baselines. Further, we ensure that the total number of env. interactions (the number of one stage or the sum of two stages) is the same across AsymDex and the baselines. See Appendix. B
for details of the grasping learning.

\textbf{AsymDex can effectively combine the acquisition and interaction phases}: We report the overall roll-out success rate of all methods for two tasks across five random seeds in Table.~\ref{tab:two stage}. We find that AsymDex significantly outperforms the other two baselines in both tasks, suggesting that combining phase decomposition with AsymDex's other two design choices (asymmetry and relative spaces) results in policies that can effectively handle both the acquisition and the interaction phases of bimanual dexterous manipulation.
The fact that \texttt{2-stage-sym} baseline outperforms the \texttt{Monolithic} baseline points to the inherent benefits of phase decomposition. Our qualitative analysis of \textit{Block in cup} task revealed that \texttt{Monolithic} policy learns to tip the cup over and push the block towards the cup. In contrast, both the two-stage policies learn more intuitive behaviors, suggesting that the phase decomposition nudges the grasping and interaction policies to learn reasonable behaviors that complement each other.

\subsection{Real-world Experiments}
We finally evaluated AsymDex and the same baselines on the following 3 real-world bimanual manipulation tasks inspired by recent bimanual manipulation works \cite{wang2024dexcap, lin2024twisting, bahety2024screwmimic, aloha2team2024aloha2enhancedlowcost, cheng2024opentelevision, ding2024bunnyvisionprorealtimebimanualdexterous, shaw2024bimanual}. 

\textbf{Hardware setup}: We use one 16-dof Allegro hand from Wonik Robotics and one 6-DoF Ability Hand from Psyonic. They are mounted on two 7-DoF Kinova Gen3 robotic arms. We used the Ability Hand for one of the hands since we only had access to one physical Allegro hand. For object pose information, we either employ an Realsense camera with AprilTag-based object tracking method or estimate it directly from the end-effector's pose, especially during occlusion.

\begin{figure}[bt]
    \centering
    \subfloat[Block in cup]{\includegraphics[width=0.15\textwidth]{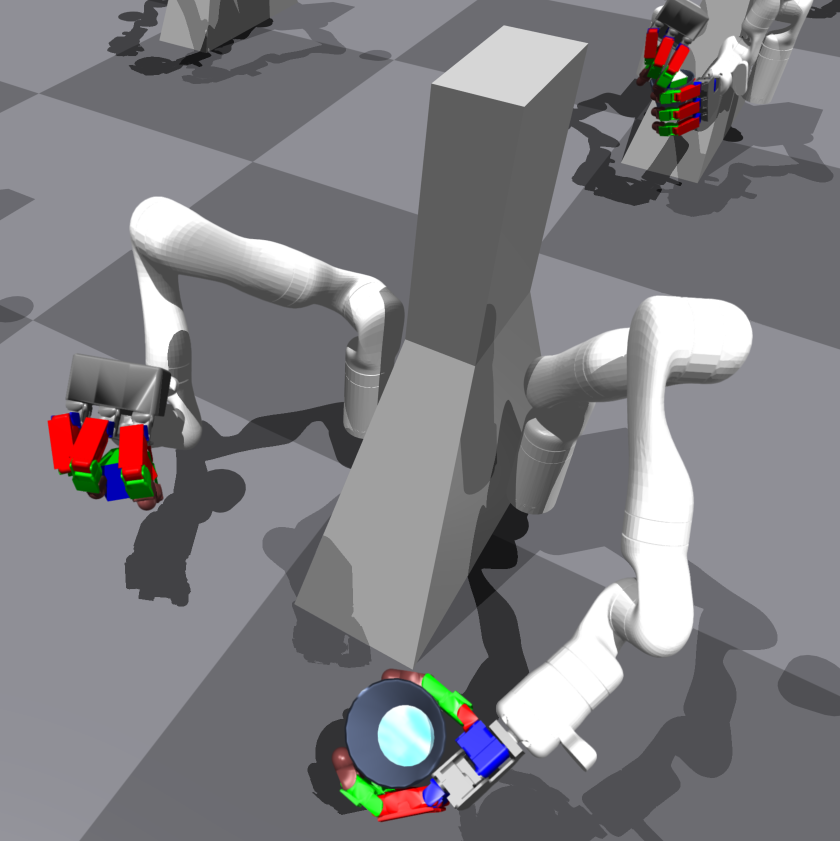}
    \includegraphics[width=0.15\textwidth]{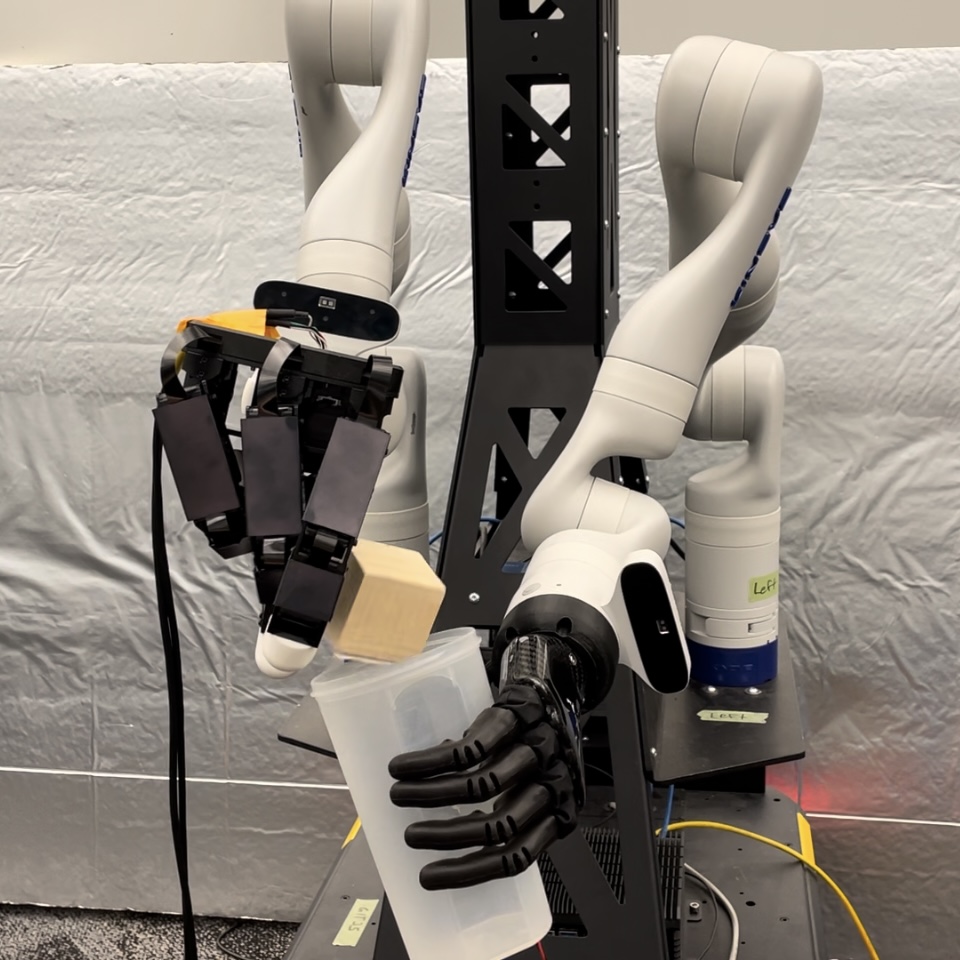}}
    \hfill
    \subfloat[Pour]{\includegraphics[width=0.15\textwidth]{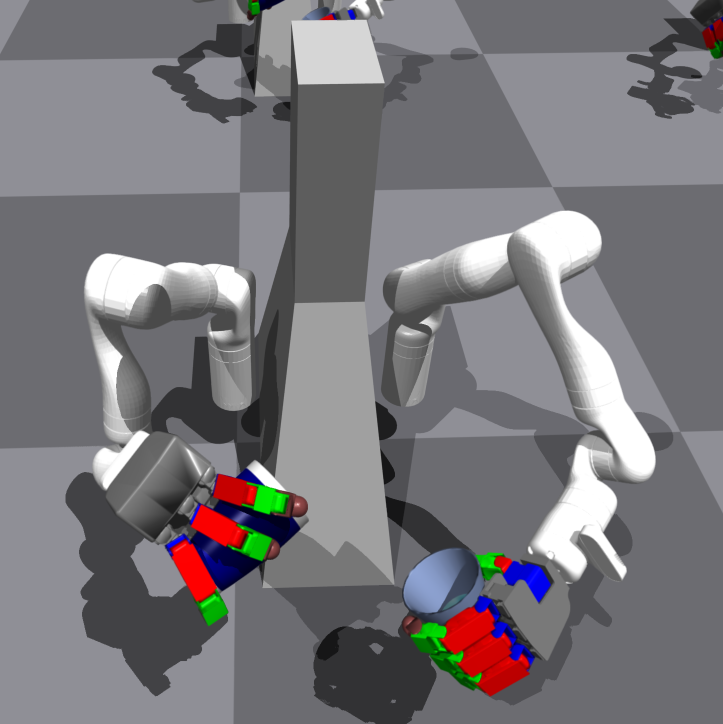}
    \includegraphics[width=0.15\textwidth]{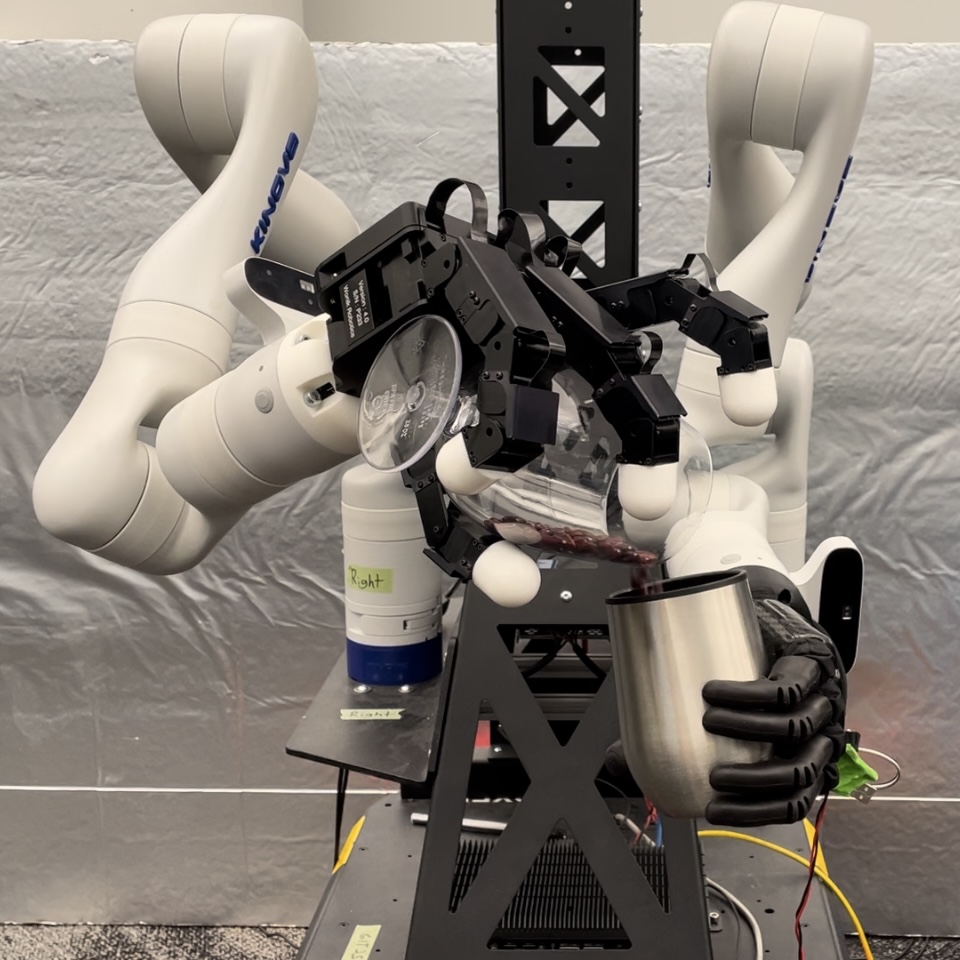}} 
    \hfill
    \subfloat[Twist lid]{\includegraphics[width=0.15\textwidth]{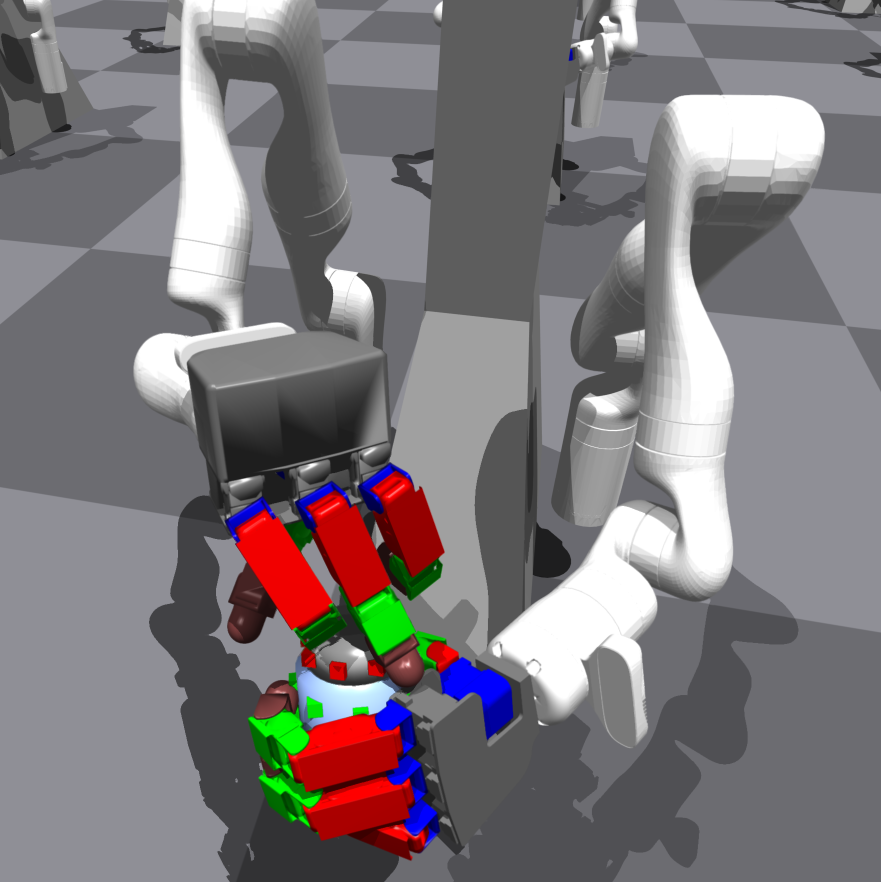}
    \includegraphics[width=0.15\textwidth]{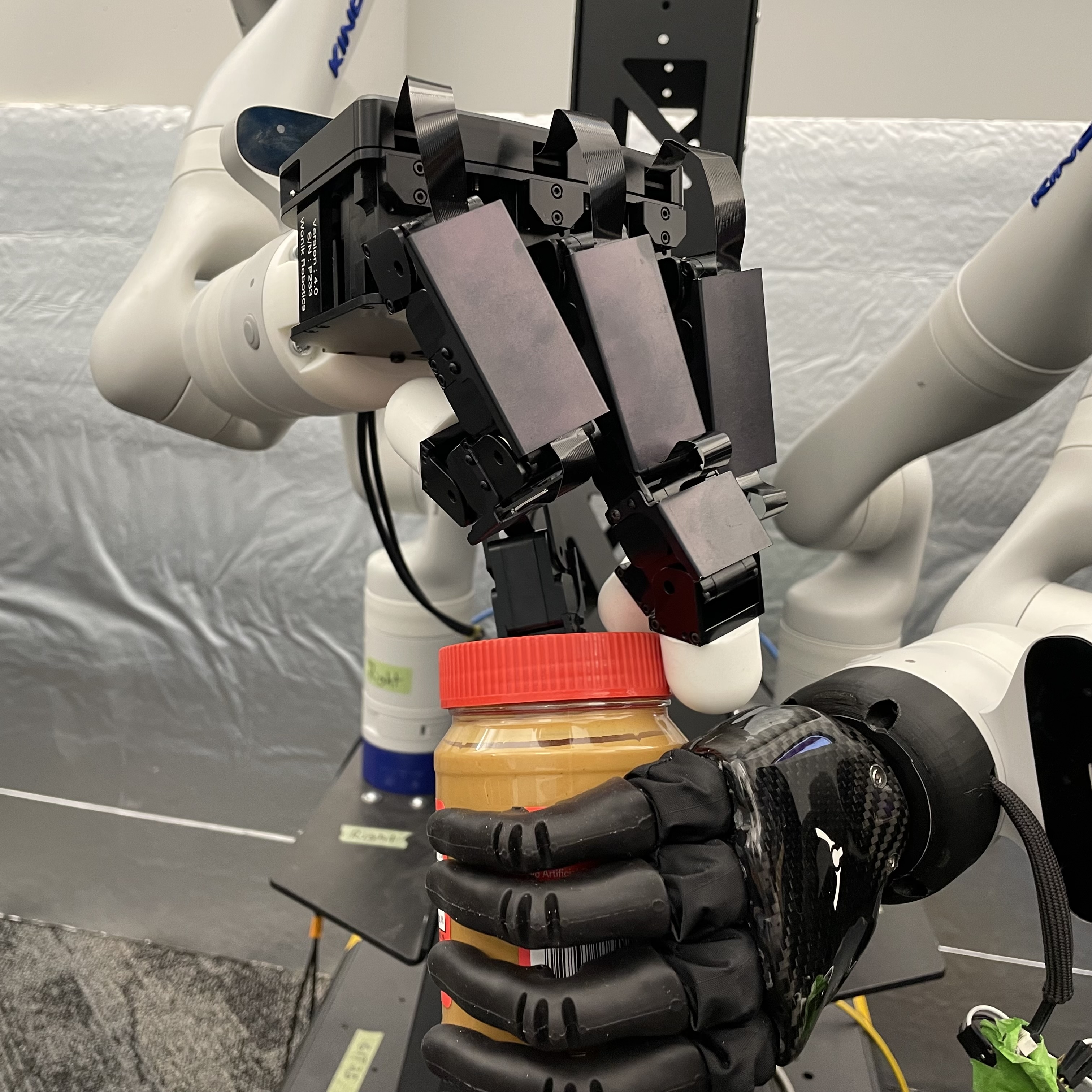}}
    \caption{We created simulation environments to match our hardware setup.}
    \label{fig:real world task images}
\end{figure}

\textbf{Real-world Tasks}: i) \textit{Block in Cup}: Place a block with sides of 5 cm inside a cylindrical cup with an internal diameter of 10 cm, ii) \textit{Pour}: Pour dry beans from one cup into another, and iii) \textit{Twist Lid}: Twist the lid off a jar while holding the jar. Snapshots are shown in Fig.~\ref{fig:real world task images}.

\textbf{Simulation Evaluation}: For each task, we first construct a simulated counterpart of our hardware setup and then train all policies in simulation.  We then compared their performance in terms of \textit{success rate} and \textit{sample efficiency}, as reported in Fig.~\ref{fig:learning curve} (b). As observed in previous experiments, AsymDex consistently and significantly outperformed all baselines in terms of both metrics.

\textbf{Sim2Real Transfer}: We deployed only the AsymDex policies on hardware since the baseline policies tended to exhibit either negligible success rates or aggressive behaviors that could damage the hardware. To enable effective sim-to-real transfer, we apply domain randomization during reinforcement learning, and details of the randomization process are provided in Appendix.~\ref{Appendix:real-world task}. 

\begin{wraptable}{r}{0.3\textwidth}
\vspace{-1em}
\resizebox{0.3\textwidth}{!}{%
\begin{tabular}{c|c|c}\toprule  
Block in cup & Pouring & Twist Lid \\\midrule
16/20 &17/20 & 18/20\\  \bottomrule
\end{tabular}
}
\caption{Real-world task success rates for AsymDex.}\label{tab:two real world}
\vspace{-1.2em}
\end{wraptable} 
\textbf{AsymDex enables zero-shot Sim2Real Transfer}: As shown in Table.~\ref{tab:two real world}, AsymDex achieves high success rates across all deployments. This once again highlights the effectiveness of AsymDex in learning robust and reliable real-world bimanual dexterous manipulation skills. For the \textit{block in cup} and \textit{pour} task, AsymDex is simply trained with observation noise component of common domain randomization technique to learn a policy that can be deployed in real world successfully.
For the most challenging \textit{twist lid} task, we added observation noise and action noise, as well as other randomization components for sim2real transfer. Though we utilized similar reward design as previous work \cite{lin2024twisting} for this task, we observed that AsymDex learns a more natural and human-like behavior compared to the previous work. Besides, unlike previous work which fixes the hand base motion \cite{grannen2023stabilize, lin2024twisting}, AsymDex is able to reposition and reorient both hand bases while two hands interact, enabling more fluent bimanual coordination. See Appendix.~\ref{Appendix:real-world task} for the detailed reward design and simulation success criterion for each task.


\section{Conclusion}
\label{sec:conclusion}
Our framework (AsymDex) is capable of learning complex asymmetric bimanual dexterous manipulation tasks via reinforcement without relying on demonstrations. We introduced and validated the need for AsymDex's two crucial ingredients: assigning asymmetric roles to the two hands, and using relative observation and action spaces. Our evaluation results reveal that the combination of these choices consistently leads to better sample efficiency and success rates across different tasks. 
	
\section{Limitations and Future Work}
\label{sec:limitations}
Our work has revealed a number of limitations and avenues for future research. First, AsymDex in its current form cannot handle certain bimanual tasks that require complex multi-finger manipulation from both hands (e.g., reorienting a heavy object, dynamic handover). Second, AsymDex does not consider the kinodynamic constraints that might result from manipulator arms. Three, behaviors produced by AsymDex are not always natural or human-like due to lack of necessary incentives.

\clearpage


\bibliography{main}  

\newpage
\appendix
\begin{center}
    {\LARGE{Appendices}}
\end{center}
\section{Training Details}
\label{Appendix:RL}
We use Proximal Policy Optimization (PPO)~\citep{schulman2017proximal} algorithm to train all policies $\pi_{\mathrm{naive}}, \pi_{\mathrm{asym}},$ and $\pi_{\mathrm{AsymDex}}$ with their corresponding value functions. Both policies and value functions are parameterized via a three-layer MLP network. The size of hidden layers for each is i) policy: (256, 256, 128), ii) value function: (512, 512, 512). The activation functions are all set as Exponential Linear Unit (ELU). We use the same PPO hyperparameters for all the baselines and AsymDex ($\gamma: 0.98, \lambda: 0.95$, clip range: 0.2, minibatch size: 8092). We use an
adaptive learning rate with KL threshold of 0.016. We train the polices on a computer with a single Nividia RTX 4090 GPU.

\begin{algorithm}[h]
\SetAlgoLined
Randomly initialize the two hand bases' poses $\xi^b_f(0)$ and $\xi^b_d(0)$, and initialize object poses $o_f(0)$ and $o_d(0)$ based on $\xi^b_f(0)$ and $\xi^b_d(0)$. Initialize policy $\pi_{\theta}$. \\
\For{$iter\in \{1, ..., \max\}$}{
Initialize replay buffer $\mathcal{B} = \varnothing$ ; \\
\For{$t\in \{1, ..., M\}$}{
\textbf{Simulate:} \\
Collect hand and object states $\xi^b_f(t)$, $\xi^b_d(t)$, $\xi^h_d(t)$, $o_f(t)$, $o_d(t)$; \\
Compute relative states $\xi^b_r(t) = \xi^b_a(t) {\circleddash} P_{f}$,  $o_r(t) = o_d(t) {\circleddash} P_{f}$; \\
Policy $\pi_{\mathrm{AsymDex}}(\hat{\xi}^b_r(t), \hat{\xi}^h_d(t)|\xi^b_r(t), \xi^h_d(t), o_r(t))$ outputs relative actions; \\
Bimanual controller (Eqn.~\ref{eqn:controller}) computes $\hat{\xi}^b_f(t)$ and $\hat{\xi}^b_d(t)$ based on $\hat{\xi}^b_r(t)$; \\
\If{Meet reset condition}{Reset environment;}
Environment physics steps with $\hat{\xi}^b_f(t),\hat{\xi}^b_d(t),\hat{\xi}^h_d(t)$; \\
\textbf{Evaluate:} \\
Compute reward $r(t)$ \\
Collect observations $(\xi^b_r(t), \xi^h_r(t), o_r(t))$, actions $(\hat{\xi}^b_r(t), \hat{\xi}^h_r(t))$, and reward $r(t)$ into buffer $\mathcal{B}$; \\
}
Update the Policy $\pi_{\mathrm{AsymDex}}$ based on  $\mathcal{B}$; \\
}
\textbf{Return:} Trained policy $\pi_{\mathrm{AsymDex}}$
\caption{AsymDex: Interaction Phase}
\label{alg:pseudo-code}
\end{algorithm} 

\section{Grasping Learning}
\label{Appendix:grasping_reward}
\paragraph{Two-stage policy} Both AsymDex (our approach) and \texttt{2-stage-sym} policy (one of the baselines in Sec.~\ref{experiment:two stage}) are two-stage policies. Therefore, they can first learn a grasping policy for the facilitating hand (or two grasping policies for facilitating hand and dominant hand respectively). Such policy $\pi_{\mathrm{grasp}}(\hat{\xi}^h_f(t)|\xi^h_f(t), \xi^b_f(t)\circleddash P_{f})$ takes in the hand joint states and the relative pose between hand and the object, and outputs the target hand joint positions to grasp the object firmly. We first provide pre-grasp annotations~\cite{dasari2023pgdm}, which allows the hands to initialize at the position close to the objects with proper joint positions. Then we script the $6D$ lifting hand base motions and design the the following rewards, which is the same across all objects.
\[
Reward = R_{rel\ pos} + R_{rel\ rot}
\]

The relative position reward $R_{rel\ pos} = (\alpha - ||x_{obj} - x_{initial}||) * \beta$, where $x_{obj}$ is the current relative position between the object and the hand, and $x_{initial}$ is the initial relative position between the object and the hand. The $\alpha, \beta \in \mathcal{R}_+$ are hyper-parameters. The relative rotation reward $R_{rel\ pos} = <u_{obj}, u_{hand}>$, where $u_{obj}$ is the object direction vector, $u_{hand}$ is the hand direction vector, and $<\cdot,\cdot>$ denotes the inner product of two vectors. We define the object direction vector and hand direction vector to be the same at the beginning of the grasping phase. Both rewards encourage the hand to keep a constant relative pose, i.e., grasping the object, during the script motion.
\paragraph{One-stage policy} Another baseline in Sec.~\ref{experiment:two stage}, i.e., the \texttt{monolithic} policy, does not incorporate the task decomposition. Therefore, it only uses the task-specific interaction rewards (see Appendix.~\ref{Appendix:task}) to learn how to complete the entire bimanual task. For a fair comparison, both hands also start at the pre-grasp poses.


\section{BiDexHands Task Design}
\label{Appendix:task}
In this section, we show the details for each BiDexHands simulation task. 
\paragraph{State Space Design}
 
For each task, the hand joint states $\xi^h_f(t)$,  $\xi^h_d(t)$ include the 24-DoF hand joint positions and the 24-DoF hand joint velocities. We use quaternions to represent rotation part of object and hand base poses. And for all policies, we also include the previous actions in the policy input. For the \textit{block in cup} task, $\xi_f^{obj}(t)$ and $\xi_d^{obj}(t)$ represent the poses of the cup and the block respectively. For the \textit{stack} task, $\xi_f^{obj}(t)$ and $\xi_d^{obj}(t)$ represent the poses of two cups. For the \textit{Bottle cap} task, $\xi_f^{obj}(t)$ and $\xi_d^{obj}(t)$ represent the poses of the bottle and the cap respectively. For the \textit{Switch} task, $\xi_f^{obj}(t)$ and $\xi_d^{obj}(t)$ represent the poses of the switch body and the button respectively. The dimensions of the observation and action spaces of each policy are shown in Table.~\ref{tab:dimensions}. It is obvious that AsymDex policy significantly reduces the state dimensions.

\begin{table}[h]
    \caption{Dimension of Observation and action spaces. For all tasks, the dimensions are identical.}
    \centering
    \begin{tabular}{c|c|c|c|c}
    \hline
    & \texttt{Sym}    & \texttt{Asym-w/o-rel} & \texttt{Rel-w/o-Asym} & \texttt{AsymDex} (ours) \\
        \hline
    \textit{Observation} & $ 176 $ &  $ 108 $ & $ 163 $  & $ 88 $ \\
        \hline
    \textit{Action} & $ 52 $ &  $ 32 $  & $ 46 $&  $ 26 $  \\
        \hline
    \end{tabular}
    \label{tab:dimensions}
\end{table}
\paragraph{Sampling Procedure}
\begin{itemize}
    \item \textit{Block in cup}: The initial position of dominant hand base is randomized: $x_d \in \mathcal{X} \sim \mathcal{U}(0.3,0.7)$, $y_d \in \mathcal{Y} \sim \mathcal{U}(-0.2, 0.0)$, $z_d \in \mathcal{Z} \sim \mathcal{U}(0.7,1.1)$. For the rotation of the dominant hand, we randomly rotate it around the axis along the arm at a random angle, $\alpha \in \mathcal{A} \sim \mathcal{U}(-1.57, 1.57)$, in radians. The block is initialized in the dominant hand. Thus its position and rotation is calculated based on the initial position and rotation of dominant hand base. The initial position of the facilitating hand base is at $[0.55, 0.6, 0.8]$.
    
    \item \textit{Stack}: The initial position of dominant hand base is randomized: $x_d \in \mathcal{X} \sim \mathcal{U}(0.3,0.7)$, $y_d \in \mathcal{Y} \sim \mathcal{U}(-0.2, 0.0)$, $z_d \in \mathcal{Z} \sim \mathcal{U}(0.7,1.1)$. For the rotation of the dominant hand, we randomly rotate it around the axis along the arm at a random angle, $\alpha \in \mathcal{A} \sim \mathcal{U}(-1.57, 1.57)$, in radians. The cup is initialized in the dominant hand. Thus its position and rotation is calculated based on the initial position and rotation of dominant hand. The initial position of the facilitating hand base is at $[0.55, 0.6, 0.8]$.
    
    \item \textit{Bottle cap}: The initial position of dominant hand base is randomized: $x_d \in \mathcal{X} \sim \mathcal{U}(0.58,0.62)$, $y_d \in \mathcal{Y} \sim \mathcal{U}(-0.21, -0.19)$, $z_d \in \mathcal{Z} \sim \mathcal{U}(0.58,0.62)$. For the rotation of the dominant hand, we randomly rotate it around the axis along the arm at a random angle, $\alpha \in \mathcal{A} \sim \mathcal{U}(-1.0, 1.0)$, in radians. The initial position of the facilitating hand base is randomized: $x_f \in \mathcal{X} \sim \mathcal{U}(0.53,0.57)$, $y_f \in \mathcal{Y} \sim \mathcal{U}(0.59, 0.61)$, $z_f \in \mathcal{Z} \sim \mathcal{U}(0.43,0.45)$. For the rotation of the facilitating hand, we randomly rotate it around the axis along the arm at a random angle, $\beta \in \mathcal{B} \sim \mathcal{U}(-0.5, 0.5)$, in radians. The bottle is initialized in the facilitating hand. Thus its position and rotation is calculated based on the initial position and rotation of facilitating hand base.
    
    \item \textit{Switch}: The initial position of dominant hand base is randomized: $x_d \in \mathcal{X} \sim \mathcal{U}(0.2,0.6)$, $y_d \in \mathcal{Y} \sim \mathcal{U}(-0.25, -0.05)$, $z_d \in \mathcal{Z} \sim \mathcal{U}(0.5,0.9)$. For the rotation of the dominant hand, we randomly rotate it around the axis along the arm at a random angle, $\alpha \in \mathcal{A} \sim \mathcal{U}(-1.0, 1.0)$, in radians. The initial position of the facilitating hand base is randomized: $x_f \in \mathcal{X} \sim \mathcal{U}(0.2,0.6)$, $y_f \in \mathcal{Y} \sim \mathcal{U}(0.05, 0.25)$, $z_f \in \mathcal{Z} \sim \mathcal{U}(0.41,0.81)$. For the rotation of the facilitating hand, we randomly rotate it around the axis along the arm at a random angle, $\beta \in \mathcal{B} \sim \mathcal{U}(-1.0, 1.0)$, in radians. The switch is initialized in the facilitating hand. Thus its position and rotation is calculated based on the initial position and rotation of facilitating hand base.
\end{itemize}
\paragraph{Success Criteria}
\begin{itemize}
    \item \textit{Block in cup} The task is considered successful if the distance of the block center and the cup center is smaller than 0.035 meters. This distance makes sure the task is only considered successful when the block is inside the cup. If the block falls on the ground or has not entered the cup within a certain time step, the task is considered failed.
    \item \textit{Stack} The task is considered successful if the distance between the cup centers is smaller than 0.02 meters. If either cup falls on the ground or has not been stacked within a certain time step, the task is considered failed.
    \item \textit{Bottle cap} The task is considered successful if the cap is taken off from its original position 0.05 meters away within a time duration, and is considered failed otherwise.
    \item \textit{Switch} The button and the switch body are connected by a revolute joint ranging from 0 to 0.5585 rads. The task is considered successful if the button is pressed and rotated 0.3585 rads within a time duration, and is considered failed otherwise.
\end{itemize}
\paragraph{Reward Design}
The reward design is similar across all tasks:
\[ Reward = \alpha_1 R_{hand\ distance} + \alpha_2 R_{progress} + \alpha_3 R_{action\ penalty} + \alpha_4 R_{success\ bonus}
\]
For each task, $R_{action\ penalty} = -||a(t)||^2$, and the $R_{success\ bonus}$ is the task success reward. $R_{hand\ distance}$ and $ R_{progress}$ are slightly different for each tasks.
\begin{itemize}
    \item \textit{Block in cup}: 
 $R_{hand\ distance} = e^{-||x_{palm} - x_{cup\ mouth}||}$, where $x_{palm}$ is the dominant hand palm position, and $x_{cup\ mouth}$ is the position of the cup mouth. $R_{progress} = -||x_{cup} - x_{block}||$, where $x_{cup}$ is the position of the cup, and $x_{block}$ is the position of the block.

\item \textit{Stack}: $R_{hand\ distance} = e^{-||x_{palm} - x_{cup\ mouth}||}$, where $x_{palm}$ is the dominant hand palm position, and $x_{cup\ mouth}$ is the position of the cup mouth, which is grasped by the facilitating hand. $R_{progress} = -||x_{cupd} - x_{cupf}||$, where $x_{cupf}$ is the position of the cup grasped by the facilitating hand, and $x_{cupd}$ is the position of the cup grasped by the dominant hand.

\item \textit{Bottle cap}: $R_{hand\ distance} = (1 - (||x_{index} - x_{cap}|| + ||x_{thumb} - x_{cap}||))^3$, where $x_{index}$ and $x_{thumb}$ are the tip position of index finger and thumb respectively, and $x_{cap}$ is the position of the bottle cap. $R_{progress} = ||x_{cap} - x_{bottle\ top}||$, where $x_{cap}$ is the position of the cap, and $x_{bottle\ top}$ is the position of the top of the bottle.
    
\item \textit{Switch}: $R_{hand\ distance} = (1 - (||x_{index} - x_{button}|| + ||x_{thumb} - x_{button}||))^3$, where $x_{index}$ and $x_{thumb}$ are the tip position of index finger and thumb respectively, and $x_{button}$ is the position of the button. $R_{progress} = 2 * \theta_{button}$, where $\theta_{button}$ is the rotated angle of the joint that connects the button and the switch body.
\end{itemize}

\section{Real-world Task Design}
\label{Appendix:real-world task}

In this section, we present the details for each real-world task.

\paragraph{State Space Design}

For each task, the hand joint states $\xi^h_f(t)$,  $\xi^h_d(t)$ include the 16-DoF hand joint positions. We use quaternions to represent rotation part of object and hand base poses. For the \textit{block in cup} task, $\xi_f^{obj}(t)$ and $\xi_d^{obj}(t)$ represent the poses of the cup and the block respectively. For the \textit{pour} task, $\xi_f^{obj}(t)$ and $\xi_d^{obj}(t)$ represent the poses of two cups. For the \textit{twist lid} task, $\xi_f^{obj}(t)$ and $\xi_d^{obj}(t)$ represent the poses of the jar and the lid respectively. The dimensions of the observation and action spaces of each policy are shown in Table.~\ref{tab:real-world dimensions}. AsymDex policy significantly reduces the state dimensions in real-world tasks consistently.

\begin{table}[h]
    \caption{Dimension of Observation and action spaces. For all tasks, the dimensions are identical.}
    \centering
    \begin{tabular}{c|c|c|c|c}
    \hline
    & \texttt{Sym}    & \texttt{Asym-w/o-rel} & \texttt{Rel-w/o-Asym} & \texttt{AsymDex} (ours) \\
        \hline
    \textit{Observation} & $ 60 $ &  $ 44 $ & $ 53 $  & $ 30 $ \\
        \hline
    \textit{Action} & $ 44 $ &  $ 28 $  & $ 38 $&  $ 22 $  \\
        \hline
    \end{tabular}
    \label{tab:real-world dimensions}
\end{table}

\paragraph{Sampling Procedure}

The corresponding simulation environments of real-world tasks include two arms and two hands attached to them. Hence, we randomize the initial position of objects and hand poses by randomizing the initial joint angles of two Kinova arms. Then we initialize the poses of objects according to the initial hand poses.

\begin{itemize}
    \item \textit{Block in cup}: The default initial joint angles of the 7-DoF Kinova arms attached to the dominant hand are $[0.0, 0.8, 0, \pi / 2 + 0.5, 0, -1.3, -\pi / 2]$. We randomized each angle by adding a $\delta \theta_i$: $\delta \theta_i \in \Theta_i \sim \mathcal{U}(-0.1, 0.1),\quad \forall i \in \{1, \dots, 4\}$;   $\delta \theta_i \in \Theta_i \sim \mathcal{U}(-0.2, 0.2),\quad \forall i \in \{5, 6\}$; $\delta \theta_7 \in \Theta_7 \sim \mathcal{U}(-0.3, 0.3)$. The default initial joint angles of the 7-DoF Kinova arms attached to the facilitating hand are $[-0.0, 0.8, 0, \pi / 2 + 0.5, 0, -1.3, 0]$. We randomized each angle by adding a $\delta \theta_i$: $\delta \theta_i \in \Theta_i \sim \mathcal{U}(-0.0, 0.0),\quad \forall i \in \{1, \dots, 3\}$;   $\delta \theta_i \in \Theta_i \sim \mathcal{U}(-0.1, 0.1),\quad \forall i \in \{4, 7\}$.

    
    \item \textit{Pour}: The default initial joint angles of the 7-DoF Kinova arms attached to the dominant hand are $[0.2, 0.8, 0, \pi / 2 + 0.5, 0, -1.3, -\pi / 2]$. We randomized each angle by adding a $\delta \theta_i$: $\delta \theta_i \in \Theta_i \sim \mathcal{U}(-0.1, 0.1),\quad \forall i \in \{1, \dots, 4\}$;   $\delta \theta_i \in \Theta_i \sim \mathcal{U}(-0.2, 0.2),\quad \forall i \in \{5, 6\}$; $\delta \theta_7 \in \Theta_7 \sim \mathcal{U}(-1.1, 0.6)$. The default initial joint angles of the 7-DoF Kinova arms attached to the facilitating hand are $[-0.2, 0.8, 0, \pi / 2 + 0.5, 0, -1.3, 0]$. We randomized each angle by adding a $\delta \theta_i$: $\delta \theta_i \in \Theta_i \sim \mathcal{U}(-0.0, 0.0),\quad \forall i \in \{1, \dots, 3\}$;   $\delta \theta_i \in \Theta_i \sim \mathcal{U}(-0.1, 0.1),\quad \forall i \in \{4, 7\}$. 
    
    \item \textit{Twist lid}: The default initial joint angles of the 7-DoF Kinova arms attached to the dominant hand are $[0.1, 0.5, 0, \pi / 2 + 0.55, 0.75, -1.5, -\pi / 2 - 0.2]$. We randomized each angle by adding a $\delta \theta_i$: $\delta \theta_i \in \Theta_i \sim \mathcal{U}(-0.0, 0.0),\quad \forall i \in \{1, \dots, 4\}$;  $\delta \theta_5 \in \Theta_5 \sim \mathcal{U}(-0.05, 0.05)$; $\delta \theta_6 \in \Theta_6 \sim \mathcal{U}(-0.2, 0.0)$; $\delta \theta_7 \in \Theta_7 \sim \mathcal{U}(-0.1, 0.1)$. The default initial joint angles of the 7-DoF Kinova arms attached to the facilitating hand are $[-0.1, 0.8, 0, \pi / 2 + 0.5, -0.7, -1.35, -0.1]$.
\end{itemize}

\paragraph{Success Criteria}

\begin{itemize}
    \item \textit{Block in cup} The task is considered successful if the distance of the block center and the cup center is smaller than 0.035 meters. This distance makes sure the task is only considered successful when the block is inside the cup. If the block falls on the ground or has not entered the cup within a certain time step, the task is considered failed.
    \item \textit{Pour} The task is considered successful if the distance between the rims of the two cups is less than 0.035 meters and the cup grasped by the facilitating hand is up-right. If either cup falls on the ground or the task is not success within a certain time step, the task is considered failed.
    \item \textit{Twist lid} We utilize the articulated bottle simulation of the previous work \cite{lin2024twisting}. The lid and bottle are connected by a single revolute joint. The task is considered successful if the lid is rotated over $3 \times \pi$ rad within a time duration, and is considered failed otherwise.
    
\end{itemize}

\paragraph{Reward Design}

The reward function structure is similar across all tasks:
\[ Reward = R_{task} + \alpha_2 R_{action\ penalty} + \alpha_3 R_{success\ bonus}
\]
For each task, $R_{action\ penalty} = -||a(t)||^2$, and the $R_{success\ bonus}$ is the task success bonus.
\begin{itemize}
    \item \textit{Block in cup}: $R_{task} = \beta_1 R_{hand\ dist} + \beta_2 R_{progress}$. $R_{hand\ dist} = e^{-||x_{palm} - x_{cup\ rim}||}$, where $x_{palm}$ is the dominant hand palm position, and $x_{cup\ rim}$ is the position of the cup rim. $R_{progress} = -||x_{cup} - x_{block}||$, where $x_{cup}$ is the position of the cup, and $x_{block}$ is the position of the block.

\item \textit{Pour}: $R_{task} = \beta_1 R_{hand\ dist} + \beta_2 R_{progress} + \beta_3 R_{cup\ orient}$. $R_{hand\ dist} = e^{-||x_{palm} - x_{cup\ rim}||}$, where $x_{palm}$ is the dominant hand palm position, and $x_{cup\ rim}$ is the position of the cup rim, which is grasped by the facilitating hand. $R_{progress} = -||x_{cupd} - x_{cupf}||$, where $x_{cupf}$ is the position of the cup rim grasped by the facilitating hand, and $x_{cupd}$ is the position of the cup rim grasped by the dominant hand. $R_{cup\ orient} = \mathbf{z}_{cupf} \cdot \mathbf{z}_{world}$, where $\mathbf{z}_{cupf}$ is the z-axis unit vector of the cup grasped by the facilitating hand, and $\mathbf{z}_{world}$ is the z-axis unit vector of the world frame.

\item \textit{Twist lid}: $R_{task} = \beta_1 R_{orient} + \beta_2 R_{twist} + \beta_3 R_{finger\ dist} + \beta_4 R_{hand\ dist\ penalty}$. $R_{orient} = \mathbf{z}_{bottle} \cdot \mathbf{z}_{world}$, where $\mathbf{z}_{bottle}$ is the z-axis unit vector of the bottle grasped by the facilitating hand, and $\mathbf{z}_{world}$ is the z-axis unit vector of the world frame. $R_{twist} = \theta_{lid}^t - \theta_{lid}^{t - 1} $, where $\theta_{lid}^t$ is the current bottle-lid revolute joint angle, and $\theta_{lid}^{t - 1}$ is the previous bottle-lid revolute joint angle. And $R_{finger\ dist}$ is the \textit{finger contact reward} we adopted from a previous lid twisting work~\cite{lin2024twisting}. $R_{hand\ dist\ penalty} = -min((||x_{f} - x_{d}|| - 0.1), 0.0)$, where $x_f$ and $x_d$ are the position of the facilitating hand palm and the dominant hand palm.

\end{itemize}

\paragraph{Domain Randomization}

The domain randomization details are shown in Table.~\ref{tab:domain randomization}

\begin{table}[h]
\centering
\caption{Domain Randomization Setup.}
\begin{tabular}{>{\bfseries}p{0.5\linewidth} p{0.4\linewidth}}
\toprule
Object: Friction (only for \textit{twist lid} task) & [0.5, 1.5] \\
\midrule
Hand: Friction (only for \textit{twist lid} task) & [0.5, 1.5] \\
\midrule
Object Pos Observation Noise & $+ \mathcal{N}(0, 0.02)$ \\
Hand Joint Observation Noise & $+ \mathcal{N}(0, 0.2)$ \\
Hand Pos Observation Noise & $+ \mathcal{N}(0, 0.02)$ \\
Hand Orientation Observation Noise & $+ \mathcal{N}(0, 0.05)$ \\
Action Noise (only for \textit{twist lid} task) & $+ \mathcal{N}(0, 0.1)$ \\
\bottomrule
\end{tabular}
\label{tab:domain randomization}
\end{table}

\end{document}